\documentclass{article}

\PassOptionsToPackage{numbers, compress, sort}{natbib}

\usepackage[preprint]{neurips_2026}

\usepackage[utf8]{inputenc} 
\usepackage[T1]{fontenc}    
\usepackage{hyperref}       
\usepackage{url}            
\usepackage{booktabs}       
\usepackage{amsfonts}       
\usepackage{nicefrac}       
\usepackage{microtype}      
\usepackage{xcolor}         
\usepackage{cite}
\usepackage[ruled, linesnumbered]{algorithm2e}

\usepackage{amsmath,amssymb,amsfonts}
\usepackage{url}
\usepackage{booktabs}
\usepackage{colortbl}
\usepackage{xurl}
\usepackage{marvosym}
\usepackage{caption}
\usepackage{subcaption}
\usepackage{multirow}
\usepackage{multicol}
\usepackage{threeparttable}
\usepackage{algorithmic}
\usepackage{graphicx}
\usepackage{textcomp}
\usepackage{color}
\definecolor{ForestGreen}{rgb}{0.13,0.55,0.13}

\usepackage{listings}
\usepackage{xspace}
\usepackage{wrapfig}
\usepackage{booktabs}
\usepackage{array}
\usepackage{enumitem}
\usepackage{placeins}
\usepackage{flafter}
\usepackage{tcolorbox}
\newcommand{\pmark}[1]{\,{\scriptsize\textcolor{gray}{$\downarrow$#1}}}
\tcbuselibrary{breakable,skins}
\newcommand{\method}{SI-AVPO\xspace}
\newcommand{\simulatorshort}{CUS\xspace}
\hypersetup{hidelinks}
\title{Unlocking Proactivity in Task-Oriented Dialogue}

\author{
Azure Zhang\textsuperscript{1}, Ning Gao\textsuperscript{1}, Yuqin Dai\textsuperscript{1}, Ruiyuan Wu\textsuperscript{1}, Jinpeng Wang\textsuperscript{1}, Rena Wei Gao\textsuperscript{1}\\ Bingdong Tan\textsuperscript{2}, Shuzheng Gao\textsuperscript{3}, Zongjie Li\textsuperscript{4}, Chaozheng Wang\textsuperscript{\Letter1} \\
\\
\textsuperscript{1}Keeta AI, Meituan
\textsuperscript{2}Independent Researcher\\
\textsuperscript{3}CUHK,  \textsuperscript{4}HKUST\\
\Letter Corresponding Author
}

\begin{document}

\maketitle

\begin{abstract}
  Proactive task-oriented dialogue (TOD), such as outbound sales, demands a persuasive agent that actively probes the user's concerns and steers the conversation toward acceptance within a bounded number of turns. Yet post-trained LLMs are inherently conservative, and reward-shaping RL (e.g., GRPO) struggles since it only re-weights what an already passive policy samples. We show that conditioning on the user's latent concerns unlocks proactive capability that no amount of sampling can undermine, establishing these concerns as a pivotal training-time signal. To operationalize this finding, we build the \textbf{Cognitive User Simulator}, which models each user as a stratified persona comprising observable external traits and hidden internal concerns. The simulator produces faithful and diverse interactions, while emitting per-turn state dynamics that track persuasion progress. We then introduce \textbf{Simulator-Induced Asymmetric-View Policy Optimization}, which converts the modeled concerns and the simulation state transition into complementary training objectives: (1) \emph{Asymmetric On-Policy Self-Distillation} that transfers concern-aware behavior from a privileged view of the same policy into its deployable, conversation-only view; and (2) \emph{State-Transition Policy Refinement} where the final decision provides trajectory-level advantage while synchronous state transitions refine turn-level credit direction. Across two real-world food-delivery benchmarks (\emph{merchant} and \emph{courier} outbound recruitment), our method matches or surpasses leading proprietary LLMs, outperforms strong RL baselines, and generalizes across various LLM-based user simulators.
\end{abstract}

\section{Introduction}\label{sec:intro}

Large Language Models (LLMs) have demonstrated strong conversational capabilities, leading interests in applying them to task-oriented dialogue (TOD) \citep{yao2024tau, barres2025tau,budzianowski2018multiwoz,bocklisch2024task}. Among TOD scenarios, \emph{proactive} settings such as outbound sales are especially challenging compared with conventional reactive ones, where agents respond to users' explicit and direct requests \citep{yao2024tau}: the agent must initiate or sustain the interaction, push the dialogue toward a predefined goal, and persuade users to take a target action \citep{dai2026sead}. These requirements impose two core demands beyond fluent response generation: agents must \textbf{identify users' inherent concerns and implicit needs} and \textbf{proactively steer the conversation toward acceptance}.

\begin{figure}[!htbp]
  \centering
  \includegraphics[width=.53\linewidth]{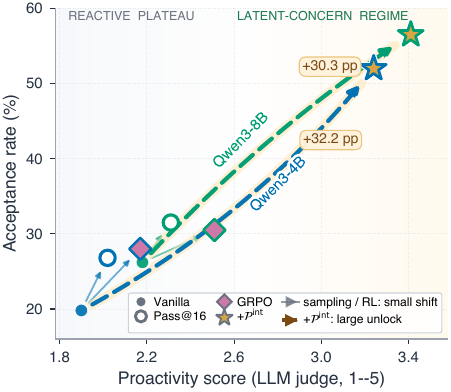}
  \caption{\textbf{Pilot study.} Latent concerns move agents from the reactive plateau to the high-proactivity/high-acceptance regime, whereas sampling and GRPO provide only small shifts.}
  \label{fig:motivating}
\end{figure}

However, these demands expose two structural gaps that existing approaches cannot bridge.
\textbf{Gap1: Current user simulators lack modeling of the implicit concerns.}
Existing simulators typically model only surface traits (e.g., tone, politeness)~\citep{zhang2026userlm, gao2026reinforcing, dai2026sead}, which makes simulated users differ mainly in style, causing dialogues to collapse into homogeneous trajectories with limited behavioral diversity. Crucially, such surface traits do not capture the \emph{latent concerns} that actually drive user decisions, making simulator decisions weakly grounded and noisy.
\textbf{Gap2: LLMs are inherently reactive.} 
Standard alignment optimizes models to \emph{respond} helpfully rather than to \emph{drive} the conversation, leaving them ill-suited for the initiative-taking required by proactive TOD. 
Mitigating this gap with current post-training formats proves equally difficult: SFT~\citep{zhou2023lima,gudibande2024false,chu2025sft} lacks high-quality data, merely mimicking surface utterances; RL methods~\citep{yu2025dapo, shao2024deepseekmath, gao2026reinforcing, dai2026sead} struggle to credit the specific turns where persuasion occurs, whether relying on sparse outcome rewards or LLM judges that score generic dialogue quality rather than actual progress toward acceptance.
These two gaps meet at the same training intersection for LLMs: the user's latent decision process. Without modeling the concerns that guide user decisions, simulators cannot provide grounded decisions, and training methods cannot identify which proactive turns actually resolve user resistance.

To further examine whether this missing variable limits current LLM policies, we conduct a pilot study to evaluate the capability and proactivity of LLMs in proactive TOD. Figure~\ref{fig:motivating} summarizes the results in the (Proactivity, Acceptance Rate) plane. Vanilla models (e.g., Qwen3-4B/8B) anchor at low operating points, and test-time scaling with Pass@16 leaves them clustered nearby, suggesting that more sampling alone cannot elicit the required persuasive behaviors. Reward-shaping RL (e.g., GRPO) consequently exhibits a similar ceiling, as it can only re-weight behaviors already sampled by the policy. By contrast, conditioning the same agents on users' latent concerns yields substantial gains in both proactivity and acceptance (dashed arrows). This confirms that the user's latent concerns are the critical missing variable: once provided, they unlock proactive capability that no amount of sampling or reward shaping can recover, establishing latent concerns not merely as a simulator detail but as a pivotal training-time signal for learning proactive policies.


Guided by this finding, we propose a co-designed framework that models latent concerns to build a faithful training environment and then converts the resulting information into complementary learning objectives.
The \textbf{Cognitive User Simulator (CUS)} models each user as a stratified persona comprising observable external traits, hidden internal concerns $\mathcal{P}^{\mathrm{int}}$, and synchronous state dynamics. By explicitly modeling the latent concerns that anchor user decisions, CUS produces behaviorally diverse and faithful interactions, providing a high-fidelity training environment that surface-trait-only simulators cannot offer. It further emits simulation feedback---the final accept/reject outcome and per-turn internal state transitions---that records how user receptiveness changes in each turn.
We then introduce \textbf{Simulator-Induced Asymmetric-View Policy Optimization} to convert the modeled concerns and the simulation state into complementary training objectives. (1) \emph{Asymmetric On-Policy Self-Distillation (AOPD)} evaluates a \emph{single} policy under two asymmetric information views: a privileged view that additionally conditions on $\mathcal{P}^{\mathrm{int}}$, and a deployable view that observes only the dialogue history. The privileged view's concern-aware behavior is progressively distilled into the deployable one, requiring no separate teacher model and enabling the agent to implicitly infer and address latent concerns at inference time. (2) \emph{State-Transition Policy Refinement (STPR)} complements this with evaluative policy-gradient signals: the final decision establishes trajectory-level advantage while per-turn state transitions refine step-level credit direction---all derived from CUS without external reward models or judges.

Extensive experiments validate both components. Human evaluation confirms that CUS produces substantially more diverse and truthful dialogue topics and more grounded decisions than prior LLM simulators. Built on CUS, our trained policy achieves higher task success on real-world outbound-call benchmarks, and is remarkably more persuasive than RL baselines and current flagship LLMs, evaluated by both powerful LLM-as-judge and human evaluators.

We make the following contributions:
\begin{itemize}[leftmargin=1.2em,itemsep=2pt,topsep=2pt]
  \item We propose Cognitive User Simulator (CUS), which models each user as a stratified persona with hidden internal concerns $\mathcal{P}^{\mathrm{int}}$ and synchronous state dynamics, calibrated on 100K+ real-world outbound-call logs, producing faithful and behaviorally diverse interactions.
  \item We introduce Simulator-Induced Asymmetric-View Policy Optimization, which converts the modeled concerns and simulation state from CUS into complementary training objectives via Asymmetric On-Policy Self-Distillation and State-Transition Policy Refinement, with no external reward model, judge, or another teacher model.
  \item Extensive experiments on two real-world food-delivery proactive-TOD benchmarks (\emph{merchant} and \emph{courier} outbound recruitment) demonstrate that our method matches or surpasses leading flagship LLMs (e.g., Claude-Sonnet-4.5, GLM-5.1) and outperforms strong RL baselines, while remaining robust across other LLM-based user simulators.
\end{itemize}

\section{Related Work}
\label{sec:related_work}

Task-oriented dialogue (TOD) has evolved from sequence-to-sequence architectures~\citep{vinyals2015neural, wen2015semantically} to LLM-driven paradigms that leverage in-context reasoning and instruction following~\citep{wang2024survey, yao2023react}, with reinforcement learning increasingly applied for policy optimization in multi-turn settings~\citep{yu2023krls, wan2025enhancing, gao2026reinforcing}. On the post-training side, alignment has progressed from RLHF/DPO~\citep{ouyang2022training, rafailov2023direct} through value-free alternatives like GRPO~\citep{shao2024deepseekmath, yu2025dapo}, multi-turn RL~\citep{zhou2024archer, zhou2025sweet}, and distillation-based methods such as OPD~\citep{agarwal2024policy} and OPSD~\citep{zhao2026self}. However, existing TOD methods lack proactive, multi-dimensional user modeling, and current post-training recipes struggle to internalize latent user state for persuasive policy learning. We address both gaps by coupling concern-grounded user simulation with asymmetric-view policy optimization.


\section{Methodology}
\label{sec:method}

\begin{figure}[!t]
\centering
\includegraphics[width=.94\linewidth]{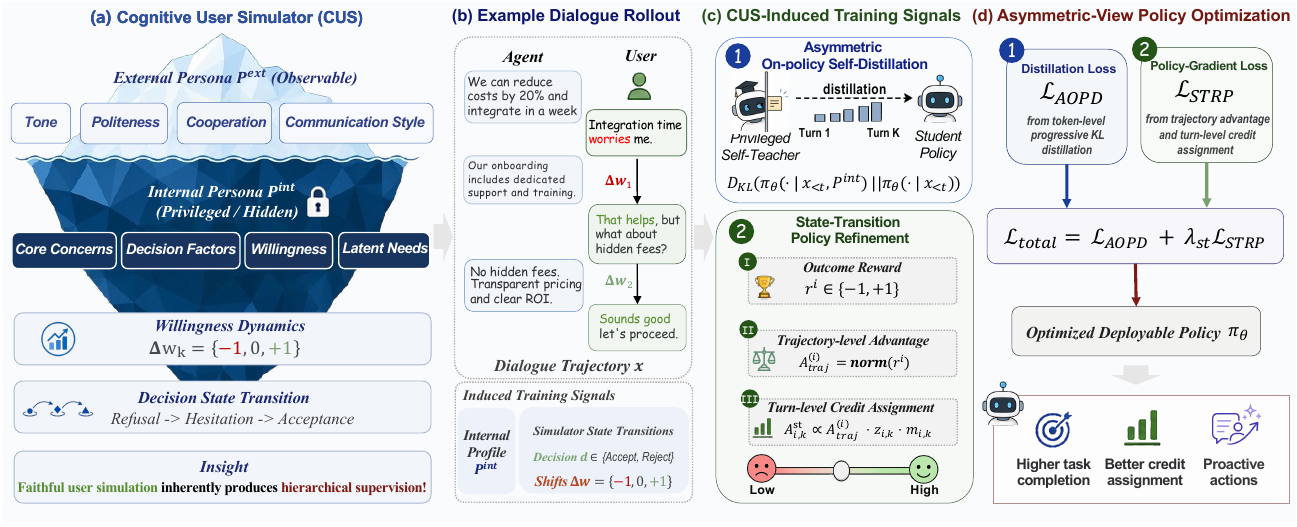}
\caption{Overview of our framework.}
\label{fig:framework}
\end{figure}

Motivated by the finding that latent concerns are the critical missing variable for proactive TOD (\S\ref{sec:intro}), we build a framework that models these concerns inside the simulator and converts them into training signals as shown in Fig. \ref{fig:framework}:
(1) The \textbf{Cognitive User Simulator (CUS)} (\S\ref{sec:cus}) represents each user with observable external traits, hidden internal concerns, and concern-driven state dynamics, yielding both a faithful training environment and two categories of natural byproduct: the privileged internal persona $\mathcal{P}^{\mathrm{int}}$ and simulation feedback (final decision and per-turn state transitions).
(2) \textbf{Simulator-Induced Asymmetric-View Policy Optimization (SI-AVPO)} (\S\ref{sec:aopd}) then converts these signals into complementary training objectives: \emph{Asymmetric On-Policy Self-Distillation (AOPD)} evaluates the same policy under a privileged view that additionally conditions on $\mathcal{P}^{\mathrm{int}}$ and progressively distills the resulting concern-aware behavior into the deployable, dialogue-only view; \emph{State-Transition Policy Refinement (STPR)} complements AOPD with policy-gradient credit, where the final decision establishes trajectory-level advantage while per-turn state transitions refine step-level credit direction---all derived from CUS without any external reward model or judge.

\subsection{Problem Definition}
\label{sec:problem}

We formalize proactive task-oriented dialogue as an episodic, undiscounted, turn-level Markov Decision Process $\mathcal{M} = \{\mathcal{S}, \mathcal{A}, P, R\}$.
Each episode samples a user profile $\mathcal{P}$ that parameterizes the transition dynamics; the detailed design are described in \S\ref{sec:cus}.
The profile governs all user-side dialogue behavior and decisions but is never exposed to the agent.

\noindent\textbf{States and actions.}\quad
The state space $\mathcal{S}$ consists of variable-length dialogue histories.
The initial state $s_0 = [x_0]$ contains only the system instruction $x_0$ (task description); the agent produces the opening utterance $y_0 \sim \pi_\theta(\cdot \mid s_0)$ \emph{before} any user response.
At subsequent turns, the state $s_k = [x_0, y_0, u_1, y_1, \dots, u_{k-1}, y_{k-1}, u_k]$ concatenates the full history up to the user's $k$-th utterance $u_k$, and the agent responds with $y_k \sim \pi_\theta(\cdot \mid s_k)$.
The action space $\mathcal{A} = \mathcal{V}^{*}$ comprises sequences over the token vocabulary.
Given $s_k$ and agent response $y_k$, the user simulator generates $u_{k+1}$ conditioned on $\mathcal{P}$ and the history, yielding $s_{k+1} = [s_k,\, y_k,\, u_{k+1}]$; the resulting transition is denoted $P_{\mathcal{P}}(s_{k+1} \mid s_k, y_k)$.

\noindent\textbf{Reward and objective.}\quad
The episode runs for at most $K$ turns and terminates with a binary user decision $d \in \{\textsc{Accept}, \textsc{Reject}\}$.
A terminal reward $r \in \{+1, -1\}$ is assigned for acceptance and rejection, respectively; the agent's objective is to maximize the expected reward:
\begin{equation}
    \max_{\pi_\theta}\;\mathbb{E}_{\mathcal{P},\; \pi_\theta}\bigl[r\bigr].
    \label{eq:objective}
\end{equation}

\subsection{Cognitive User Simulator}
\label{sec:cus}

To provide a faithful environment that produces behaviorally diverse and grounded interactions, we propose a stratified user simulator framework that models the latent concerns that anchor user decisions and couples state dynamics to concern progress.

\noindent\textbf{Three-Layer Persona Framework.}\quad
We instantiate \simulatorshort{} for proactive TOD tasks in collaboration with the operations team of a major food-delivery platform; the persona schema and the concern bank described below are co-designed with domain specialists, while a corpus of $100\mathrm{K}{+}$ outbound call records provides a stylistic and behavioral prior that calibrates the simulator's language, pacing, and reaction patterns.
Each user is represented as a stratified profile $\mathcal{P} = \bigl(\mathcal{P}^{\mathrm{bg}},\; \mathcal{P}^{\mathrm{ext}},\; \mathcal{P}^{\mathrm{int}}\bigr)$
whose three layers play distinct functional roles.
The (1) \emph{background} $\mathcal{P}^{\mathrm{bg}}$ specifies the situational frame at call onset (e.g., demographic attributes) and is naturally exposed through the call.
The (2) \emph{external persona} $\mathcal{P}^{\mathrm{ext}}$ governs surface behavior along five observable axes (time pressure, courtesy, communication style, cooperation tendency, technical familiarity) and shapes the texture of every user utterance the agent receives.

The (3) \emph{internal persona} $\mathcal{P}^{\mathrm{int}}$ encodes the user's latent concerns across financial, psychological, and operational facets.
Each $\mathcal{P}^{\mathrm{int}}$ dimension is grounded by a curated \emph{concern bank}: every concern contains a natural-language resistance point, unlock conditions that specify when it can be resolved, and anti-patterns that penalize harmful or over-promising agent behavior.
$\mathcal{P}^{\mathrm{int}}$ directly governs both the per-turn state dynamics and the terminal decision: willingness can only rise meaningfully through concern resolution, and acceptance is gated by sufficient concern coverage---ensuring that the simulator's decisions reflect genuine concern progress rather than free-form LLM judgments.

\noindent\textbf{Simulation State Transitions.}\quad
Beyond the static persona, \simulatorshort{} maintains a scalar \emph{willingness} state that tracks how close the user is to acceptance and is updated turn by turn.
At each agent turn $k$, the simulator emits a textual user response $u_k$, a discrete action $a_k\!\in\!\{\textsc{chat},\textsc{hangup}\}$, a signed willingness shift $\Delta w_k$, and per-concern state transitions $\Delta\mathcal{C}_k$ that track each concern through monotone states (\textsc{unresolved} $\to$ \textsc{partially-addressed} $\to$ \textsc{resolved}).
Importantly, $\Delta w_k$ is not a free-form LLM score: it is rule-bounded by concern progression, so positive shifts require actual concern resolution and anti-pattern hits induce penalties.
The episode terminates with a binary decision $d \in \{\textsc{Accept}, \textsc{Reject}\}$, gated by willingness and concern coverage.

\noindent\textbf{Summary.}\quad
This design yields one privileged state signal and two natural state-transition signals, which are all available without any external annotation:
\begin{equation}
    \bigl(
    \underbrace{\mathcal{P}^{\mathrm{int}}}_{\text{privileged internal state}},
    \quad
    \underbrace{d,\{\Delta w_k\}_{k=1}^{K}}_{\text{simulation state-transitions}}
    \bigr).
    \label{eq:signals}
\end{equation}
The internal state, which specifies the latent concerns invisible to the deployed agent, supports asymmetric self-distillation; the final decision and per-turn willingness shifts provide trajectory- and turn-level credit assignment.
The next section composes these signals into the training objective.

\subsection{Simulator-Induced Asymmetric-View Policy Optimization}
\label{sec:aopd}

To utilize the user's invisible latent decision process induced by CUS and derive training signals, we propose \textbf{Simulator-Induced Asymmetric-View Policy Optimization (SI-AVPO)}.

\noindent\textbf{Overall Objective.}\quad
 SI-AVPO has two complementary components that convert the aforementioned signals into training objectives, respectively. First, \textit{Asymmetric On-Policy Self-Distillation (AOPD)} transfers concern-aware behavior from a privileged view of the same policy into its deployable dialogue-only view. Second, \textit{State-Transition Policy Refinement (STPR)} uses the simulator's final decision and synchronous state (i.e., willingness) transitions to refine on-policy credit assignment:
\begin{equation}
    \mathcal{L}_{\mathrm{SI\text{-}AVPO}}
    = \mathcal{L}_{\mathrm{AOPD}}(\mathcal{P}^{\mathrm{int}})
    + \lambda_{\mathrm{st}}\,\mathcal{L}_{\mathrm{STPR}}(d,\{\Delta w_k\}_{k=1}^{K}).
    \label{eq:si-avpo}
\end{equation}

\noindent\textbf{Asymmetric On-Policy Self-Distillation.}\quad
AOPD aims to turn the internal persona exposed by CUS at training time into token-level supervision for a dialogue-only policy.
We define two information views of the same policy $\pi_\theta$: a \emph{privileged teacher} view whose context is augmented with the user's internal persona $\mathcal{P}^{\mathrm{int}}_i$, and a \emph{deployable student} view that observes only the task instruction and dialogue history.
For each student-generated token in an on-policy rollout, we compute the probability distributions under both views and minimize their divergence, thereby distilling concern-aware behavior from the teacher into the student:
\begin{equation}
    \mathcal{L}_{\mathrm{AOPD}}
    = \frac{1}{N}\sum_{i,t}\phi(k(t),K)
      D_{\mathrm{KL}}\!\left(
        \mathrm{sg}\!\left[\pi_\theta\!\left(\cdot \mid x_{i,t},\mathcal{P}_i^{\mathrm{int}}\right)\right]
        \;\middle\|\;
        \pi_\theta\!\left(\cdot \mid x_{i,t}\right)
      \right).
    \label{eq:aopd}
\end{equation}
Here, $N$ is the number of assistant tokens, $K$ is the maximum turns, $k(t)$ maps token $t$ to its dialogue turn, and $\phi(k(t),K)$ is a nonnegative turn-dependent distillation weight.
The stop-gradient operator $\mathrm{sg}[\cdot]$ freezes the teacher distribution as a fixed target, so gradients flow only through the student view.
Because both views share the same $\pi_\theta$, no separate teacher model is needed; at inference time the policy operates under the student view alone, without any dependence on $\mathcal{P}^{\mathrm{int}}$.


\noindent\textbf{State-Transition Policy Refinement.}\quad
While AOPD provides directive supervision that teaches the student \emph{what} a concern-aware agent would say, it leaves the question of \emph{whether} a particular attempt actually steers the user's acceptance, e.g., a turn may probe the right concern at the wrong moment that triggers user resistance.
Furthermore, since $\mathcal{P}^{\mathrm{int}}$ is unavailable at inference, perfect imitation of the privileged teacher is inherently unattainable; the student must explore and refine its own dialogue strategies under the constrained information view, yet distillation alone offers no evaluative signal to guide this exploration.

Therefore, we propose STPR to address both limitations by introducing evaluative credit grounded in the simulator's state dynamics. STPR uses CUS state transition dynamics to assign trajectory credit. 
Formally, the final decision defines a normalized trajectory advantage $A_{\mathrm{traj}}^{(i)}=\frac{r_i-\mu_r}{\sigma_r+\varepsilon_{\mathrm{norm}}}$, setting the trajectory-level update direction and scale.
Here, $r_i$ is the terminal reward for rollout $i$, $\mu_r$ and $\sigma_r$ are the mean and standard deviation of rewards within the rollout group, and $\varepsilon_{\mathrm{norm}}$ is a small positive constant for numerical stability.
The per-turn willingness shift $\Delta w_{i,k}$ then adjusts this signal: a transition that directionally aligns with the final outcome reinforces the trajectory direction, a contradictory transition flips it, and a zero transition leaves the trajectory credit unchanged.
Let $\Delta w_{\max}$ denote the maximum absolute willingness shift over all turns in the rollout group.
For turns with a nonzero willingness shift, let $m_{i,k}=|\Delta w_{i,k}|/\Delta w_{\max}$ denote the magnitude gate and $z_{i,k}=\operatorname{sgn}(A_{\mathrm{traj}}^{(i)}\Delta w_{i,k})$ denote the alignment sign. The state-transition advantage is:
\begin{equation}
A_{i,k}^{\mathrm{st}}
=
A_{\mathrm{traj}}^{(i)} z_{i,k}m_{i,k}\exp\!\left(\tau z_{i,k}\right),
\quad \Delta w_{i,k}\neq0.
\label{eq:st_adv}
\end{equation}
The magnitude gate $m_{i,k}$ measures how much the simulator state changes at turn $k$, and the temperature $\tau$ controls the strength of outcome--transition polarization: larger values amplify turns whose willingness shift aligns with the final outcome and more strongly suppress contradictory turns.
The alignment sign $z_{i,k}$ determines the credit direction: it is positive when the local willingness shift agrees with the terminal trajectory advantage and negative when it contradicts it.
The exponential factor then further amplifies aligned transitions and suppresses contradictory ones.
If $\Delta w_{i,k}=0$, we directly set $A_{i,k}^{\mathrm{st}}=A_{\mathrm{traj}}^{(i)}$, so turns without simulator state change keep the original trajectory credit.
STPR therefore anchors credit to the trajectory-level advantage while assigning aligned turns same-signed credit and contradictory turns opposite-signed credit according to the simulator's state dynamics. Broadcasting $A_{i,k}^{\mathrm{st}}$ to all assistant tokens in turn $k$ gives the clipped refinement loss
\begin{equation}
    \mathcal{L}_{\mathrm{STPR}}
    = -\frac{1}{N}\sum_{i,t}
      \min\!\left(
        \rho_{i,t}\, A_{i,k(t)}^{\mathrm{st}},\;
        \mathrm{clip}\!\left(\rho_{i,t}, 1\!-\!\varepsilon_{\mathrm{clip}}, 1\!+\!\varepsilon_{\mathrm{clip}}\right)
        A_{i,k(t)}^{\mathrm{st}}
      \right),
    \label{eq:st_pg}
\end{equation}
where $\rho_{i,t}=\pi_\theta(y_{i,t}\mid x_{i,t})/\pi_{\theta_{\mathrm{old}}}(y_{i,t}\mid x_{i,t})$. Algorithm~\ref{alg:aopd} summarizes the complete training loop.

\begin{algorithm}[!t]
\footnotesize
\SetAlgoNlRelativeSize{0}
\DontPrintSemicolon
\SetAlgoLined
\SetAlgoSkip{smallskip}
\SetInd{0.35em}{0.6em}
\SetKwInOut{Input}{Input}
\SetKwInOut{Output}{Output}
\SetKwInOut{Hyper}{Hyper}
\SetKw{KwReturn}{return}

\Input{policy $\pi_\theta$; CUS persona sampler $\mathcal{P}$; group size $G$; max turns $K$; inner epochs $E$}
\Hyper{$\varepsilon_{\rm clip},\varepsilon_{\rm norm},\phi(k,K),\tau,\lambda_{\rm st}$}
\Output{deployable policy $\pi_\theta$}

\For{\textup{each training iteration}}{
  $\theta_{\rm old}\leftarrow\theta$\;

  \tcp{\textbf{Phase 1:} collect CUS rollouts under the deployable view}
  Sample $\mathcal{P}_i\sim\mathcal{P}$\;
  \For{$i=1$ \KwTo $G$}{
    Run a $K$-turn dialogue with
    $\pi_{\theta_{\rm old}}(\cdot\mid x)$\;
    Record $\{(x_{i,t},y_{i,t},k(t))\}$, hidden profile $\mathcal{P}_i^{\rm int}$,
    shifts $\{\Delta w_{i,k}\}_{k=1}^{K}$, and reward $r_i$\;
  }

  \tcp{\textbf{Phase 2:} compute simulator-grounded STPR credit}
  $A_{\rm traj}^{(i)}\leftarrow (r_i-\mu_r)/(\sigma_r+\varepsilon_{\rm norm})\ \forall i$,\quad
  $\Delta w_{\max}\leftarrow \max_{j\in[G],\,\ell\in[K]}|\Delta w_{j,\ell}|$\;

  \ForEach{turn $(i,k)$}{
    $z_{i,k}\leftarrow \operatorname{sgn}(A_{\rm traj}^{(i)}\Delta w_{i,k})$,\quad
    $m_{i,k}\leftarrow |\Delta w_{i,k}|/\Delta w_{\max}$\;
    $\displaystyle
    A_{i,k}^{\rm st}\leftarrow
    \begin{cases}
      A_{\rm traj}^{(i)}, & \Delta w_{i,k}=0,\\
      A_{\rm traj}^{(i)} z_{i,k}m_{i,k}\exp(\tau z_{i,k}), & \Delta w_{i,k}\neq 0;
    \end{cases}$\;
    Broadcast $A_{i,k}^{\rm st}$ to all assistant tokens in turn $k$\;
  }

  \tcp{\textbf{Phases 3--4:} privileged AOPD loss and joint update}
  \For{$e=1$ \KwTo $E$ \textup{ and each minibatch } $\mathcal{B}$}{
    \ForEach{$(i,t)\in\mathcal{B}$}{
      $q_{i,t}^{\rm priv}\leftarrow
      \mathrm{sg}\!\left[\pi_\theta(\cdot\mid x_{i,t},\mathcal{P}_i^{\rm int})\right]$,\quad
      $p_{i,t}\leftarrow\pi_\theta(\cdot\mid x_{i,t})$,\quad
      $\displaystyle
      \rho_{i,t}\leftarrow
      \frac{\pi_\theta(y_{i,t}\mid x_{i,t})}
           {\pi_{\theta_{\rm old}}(y_{i,t}\mid x_{i,t})}$\;
    }

    $\displaystyle
    \mathcal{L}_{\rm AOPD}^{\mathcal{B}}\leftarrow
    \frac{1}{|\mathcal{B}|}\sum_{(i,t)\in\mathcal{B}}
    \phi(k(t),K)
    D_{\rm KL}\!\left(q_{i,t}^{\rm priv}\middle\|p_{i,t}\right)$\;

    $\displaystyle
    \mathcal{L}_{\rm STPR}^{\mathcal{B}}\leftarrow
    -\frac{1}{|\mathcal{B}|}\sum_{(i,t)\in\mathcal{B}}
    \min\!\left(
      \rho_{i,t}A_{i,k(t)}^{\rm st},
      \operatorname{clip}(\rho_{i,t},1-\varepsilon_{\rm clip},1+\varepsilon_{\rm clip})
      A_{i,k(t)}^{\rm st}
    \right)$\;

    $\displaystyle
    \theta\leftarrow\theta-\nabla_\theta
    \left(
      \mathcal{L}_{\rm AOPD}^{\mathcal{B}}
      +\lambda_{\rm st}\mathcal{L}_{\rm STPR}^{\mathcal{B}}
    \right)$\;
  }
}
\KwReturn $\pi_\theta$\;

\caption{\textbf{Simulator-Induced Asymmetric-View Policy Optimization} (\textsc{SI-AVPO}).}
\label{alg:aopd}
\end{algorithm}

\section{Experimental Setup}

\noindent\textbf{Baselines.} We compare \method against two categories of baselines. \textbf{(1) Large Models.} We select five state-of-the-art proprietary LLMs—GPT-4.1~\citep{openai2025gpt41}, DeepSeek-V3.2~\citep{liu2025deepseek}, GLM-5.1~\citep{glm51}, Kimi-K2.5~\citep{team2026kimi}, and Claude-Sonnet-4.5~\citep{claudesonnet45}—with the same task instructions to establish upper-bound references. \textbf{(2) RL-based Methods.} Starting from two foundation models, Qwen3-4B and Qwen3-8B~\citep{yang2025qwen3}, we apply several representative reinforcement learning algorithms: GRPO~\citep{shao2024deepseekmath}, PPO~\citep{schulman2017proximal}, DAPO~\citep{yu2025dapo}, and SEAD~\citep{dai2026sead}. All RL baselines are trained with the same user simulator and environment configuration for a fair comparison.


\noindent\textbf{Datasets.} We instantiate \method on two real-world proactive outbound-call tasks from our business: \emph{Merchant Promotion} (recruiting new merchants into a paid traffic campaign) and \emph{Courier Regional Bonus} (enrolling couriers into a region-specific rush-hour bonus program).  For each task we sample $10{,}000$ personas for training and a disjoint set of $200$ personas for evaluation, so test-time performance reflects generalization to unseen profile combinations rather than memorization. During persona construction, each user's specific internal concerns are sampled uniformly at random from the task-specific concern bank, with the number of concerns drawn from $\mathrm{Uniform}\{3,6\}$, yielding diverse difficulty levels across episodes. The shared external profile covers five observable dimensions, while the task-specific internal profile contains 14 dimensions/42 values/89 curated concerns for merchants and 12 dimensions/36 values/78 curated concerns for couriers; all concerns are audited to remain within the agent's authorized scope.

\noindent\textbf{Evaluation Metrics.} We measure two objective signals from the simulator's internal state transitions: the \textbf{Acceptance Rate (AR)}, the percentage of dialogues ending in user acceptance, and the \textbf{Concern Solving Rate (CSR)}, the proportion of the user's hidden concerns successfully addressed during the interaction. To assess interaction quality beyond task completion, we employ three strong third-party LLMs (Gemini-3.1-Pro~\citep{gemini31pro}, GPT-5.4~\citep{openai2025gpt54}, and Claude-Opus-4.6~\citep{claudeopus46}) as judges to score each dialogue on a 1--5 scale along three dimensions: \textbf{Communication}, \textbf{Logic}, and \textbf{Proactivity}. We average the independent judge scores for each dialogue and report means across repeated evaluation runs.

\noindent\textbf{Implementation Details.}\label{sec:implementation}
All experiments are implemented using veRL~\citep{sheng2024hybridflow}. For RL training, we use 8$\times$ NVIDIA A100 (80GB) GPUs with batch size 128, learning rate $1\times10^{-6}$, three inner optimization epochs, eight rollouts per prompt, and a maximum dialogue length of 20 interaction turns. The user simulator is served on 16$\times$ NVIDIA H20 GPUs to support high-throughput parallel rollout generation with bfloat16 SGLang inference. We use Qwen3.5-115B~\citep{qwen35} as the default user simulator during training and Claude-Sonnet-4.5 for evaluation. For evaluation, each experiment is repeated \emph{three times}. For LLM-as-judge metrics, each dialogue is scored by the three judges introduced above independently, and we report the average scores.

\section{Experiment Results}

\subsection{Main Results}
We first evaluate \method{} against two groups of baselines: representative RL algorithms built on the same Qwen backbones and flagship proprietary LLMs prompted with the same task instructions.
Table~\ref{tab:fds_performance_comparison} summarizes both task success and dialogue-quality metrics on the merchant and courier outbound-call scenes.

\begin{table}[!t]
\centering
\caption{Model performance on \textit{FoodDeliverService}. CSR and Acc are percentages; Comm., Logic, and Proact. are 1--5 judge scores. \textbf{Bold}: best overall; \underline{underline}: second best overall.}
\label{tab:fds_performance_comparison}
\scriptsize
\setlength{\tabcolsep}{2.1pt}
\renewcommand{\arraystretch}{1.06}
\definecolor{methodblue}{RGB}{220,235,252}
\definecolor{headergray}{RGB}{240,240,240}
\resizebox{\linewidth}{!}{%
\begin{tabular}{lrrrrrrrrrr}
\toprule
\multirow{3}{*}{\textbf{Method}} & \multicolumn{5}{c}{\textbf{Merchant Promotion}} & \multicolumn{5}{c}{\textbf{Courier Regional Bonus}} \\
\cmidrule(lr){2-6}\cmidrule(lr){7-11}
& \multicolumn{2}{c}{\textbf{Task}} & \multicolumn{3}{c}{\textbf{Dialogue}} & \multicolumn{2}{c}{\textbf{Task}} & \multicolumn{3}{c}{\textbf{Dialogue}} \\
\cmidrule(lr){2-3}\cmidrule(lr){4-6}\cmidrule(lr){7-8}\cmidrule(lr){9-11}
& CSR $\uparrow$ & Acc $\uparrow$ & Comm. $\uparrow$ & Logic $\uparrow$ & Proact. $\uparrow$ & CSR $\uparrow$ & Acc $\uparrow$ & Comm. $\uparrow$ & Logic $\uparrow$ & Proact. $\uparrow$ \\
\midrule
\rowcolor{headergray}
\multicolumn{11}{l}{\textit{Large Models}} \\
GPT-4.1 & 36.1 & 34.7 & 2.93 & 3.02 & 2.12 & 30.7 & 22.8 & 2.66 & 2.91 & 1.79 \\
DeepSeek-v3.2 & 45.7 & 39.5 & 3.04 & 3.13 & 2.45 & 32.6 & 28.0 & 2.54 & 3.27 & 1.96 \\
GLM-5.1 & 46.2 & 42.3 & 3.32 & 3.44 & 2.78 & 38.7 & 38.7 & 2.86 & 3.56 & 2.12 \\
Kimi-K2.5 & 49.0 & 39.8 & 3.26 & 3.51 & 2.67 & 39.0 & 38.8 & 2.79 & 3.45 & 2.21 \\
Claude-Sonnet-4.5 & 51.1 & \textbf{50.8} & \underline{3.56} & \textbf{3.76} & 3.09 & \underline{45.0} & \underline{43.2} & \underline{3.03} & \textbf{3.75} & \underline{2.67} \\
\midrule
\rowcolor{headergray}
\multicolumn{11}{l}{\textit{Qwen-3-4B}} \\
~~Base & 22.5 & 20.2 & 2.23 & 2.15 & 1.90 & 20.8 & 18.5 & 1.96 & 2.34 & 1.76 \\
~~+GRPO & 28.4 & 28.0 & 2.47 & 2.61 & 2.17 & 24.3 & 22.3 & 2.13 & 2.55 & 1.86 \\
~~+PPO & 26.5 & 22.3 & 2.36 & 2.59 & 2.22 & 22.7 & 19.8 & 2.06 & 2.44 & 1.75 \\
~~+DAPO & 29.7 & 29.5 & 2.60 & 2.74 & 2.33 & 23.8 & 26.3 & 2.25 & 2.81 & 2.10 \\
~~+SEAD & 33.7 & 32.7 & 2.95 & 3.22 & 2.58 & 30.2 & 32.5 & 2.52 & 3.21 & 2.26 \\
\rowcolor{methodblue}
~~+\method & \underline{51.9} & 46.3 & 3.25 & 3.56 & \underline{3.11} & 44.6 & 42.0 & 2.87 & 3.52 & 2.63 \\
\midrule
\rowcolor{headergray}
\multicolumn{11}{l}{\textit{Qwen-3-8B}} \\
~~Base & 30.6 & 26.2 & 2.66 & 2.48 & 2.18 & 23.5 & 22.2 & 2.35 & 2.51 & 2.00 \\
~~+GRPO & 33.2 & 30.5 & 2.69 & 2.88 & 2.51 & 29.6 & 24.7 & 2.41 & 3.03 & 2.11 \\
~~+PPO & 31.8 & 26.8 & 2.60 & 2.81 & 2.26 & 25.2 & 24.8 & 2.40 & 2.87 & 2.13 \\
~~+DAPO & 36.3 & 34.3 & 2.81 & 3.37 & 2.41 & 30.7 & 28.7 & 2.59 & 3.19 & 2.28 \\
~~+SEAD & 40.4 & 36.3 & 3.23 & 3.56 & 2.63 & 36.7 & 33.5 & 2.75 & 3.51 & 2.45 \\
\rowcolor{methodblue}
~~+\method & \textbf{52.6} & \underline{48.5} & \textbf{3.61} & \underline{3.71} & \textbf{3.18} & \textbf{45.6} & \textbf{44.3} & \textbf{3.06} & \underline{3.66} & \textbf{2.94} \\
\bottomrule
\end{tabular}%
}
\end{table}

Among the RL baselines, general-purpose algorithms (GRPO, DAPO, PPO) yield modest improvements over the base model in task scores and dialogue quality, yet their gains on the proactivity remain limited. For example, on Qwen3-8B, DAPO raises Proact.\ from 2.18 to only 2.41 on the merchant scene.
This pattern suggests that outcome-driven RL can refine surface-level communication skills and help the agent better address user concerns once they are raised, but it struggles to induce genuinely proactive behaviors such as anticipating latent objections or steering the conversation toward unspoken needs.
SEAD advances further through its curriculum-based self-evolution, achieving stronger performance.
Nevertheless, \method{} consistently and substantially outperforms all RL baselines across both scenes and both backbone sizes.
On Qwen3-8B, \method{} significantly surpasses the best RL baseline (SEAD) by +12.2\% CSR, +12.2\% Acc, and +0.55 Proact.\ on the merchant scene, and by +8.9\% CSR, +10.8\% Acc, and +0.49 Proact.\ on the courier scene (with p-value less than $10^{-5}$).
The improvement in proactivity indicates that our SI-AVPO provides direct supervision for initiative-taking behaviors that conventional RL rarely discovers on its own: actively probing hidden concerns, adapting to user hesitation, and steering conversations toward decisions rather than passively responding.

Notably, \method{} outperforms four of the five flagship LLMs, including GPT-4.1, DeepSeek-v3.2, GLM-5.1, and Kimi-K2.5, on most metrics across both scenes, even with the smaller Qwen3-4B backbone.
With Qwen3-8B, \method{} further rivals the strongest flagship and proprietary model, Claude-Sonnet-4.5, surpassing it on CSR (52.6\% vs.\ 51.1\%), Proact.\ (3.18 vs.\ 3.09) on the merchant scene, and on both CSR and Acc on the courier scene, while trailing only marginally on Acc and Logic for merchants.
Since flagship models of this scale are impractical for real-time outbound calling due to latency, cost, and data-governance constraints, the key takeaway is that a deployable-size policy trained with SI-AVPO can match or exceed much larger general-purpose models in proactive task-oriented dialogue.

\subsection{Ablation Study}

Table~\ref{tab:ablation} presents the ablation results, from which we draw the following observations.

\begin{table}[!tbp]
\centering
\caption{Ablation on \textit{Merchant} (Qwen3-4B).}
\label{tab:ablation}
\footnotesize
\setlength{\tabcolsep}{4pt}
\definecolor{fullblue}{RGB}{220,235,252}
\begin{tabular}{@{}lccccc@{}}
\toprule
\textbf{Variant} & \textbf{CSR}{\footnotesize(\%)} & \textbf{Acc}{\footnotesize(\%)} & \textbf{Comm.} & \textbf{Logic} & \textbf{Proact.} \\
\midrule
\rowcolor{fullblue}
\method{} (Full)                     & \textbf{51.9} & \textbf{46.3} & \textbf{3.25} & \textbf{3.56} & 3.11 \\
\addlinespace[2pt]
\multicolumn{6}{@{}l}{\textit{\footnotesize Removing AOPD components}} \\
~~w/o AOPD {\footnotesize($\to$STPR)}             & 35.4\pmark{16.5} & 34.5\pmark{11.8} & 2.81 & 3.01 & 2.35 \\
~~w/o AOPD\,+\,Turn {\footnotesize($\to$GRPO)}    & 28.4\pmark{23.5} & 28.0\pmark{18.3} & 2.47 & 2.61 & 2.17 \\
\addlinespace[2pt]
\multicolumn{6}{@{}l}{\textit{\footnotesize Removing STPR components}} \\
~~w/o Turn-Level Reward              & 46.7\pmark{5.2}  & 42.2\pmark{4.1}  & 3.01 & 3.21 & 3.14 \\
~~w/o STPR {\footnotesize($\to$AOPD)}             & 44.6\pmark{7.3}  & 38.7\pmark{7.6}  & 2.85 & 2.98 & \textbf{3.23} \\
~~w/o STPR\,+\,Asym {\footnotesize($\to$OPD)}     & 37.6\pmark{14.3} & 32.8\pmark{13.5} & 2.66 & 3.01 & 2.39 \\
\bottomrule
\end{tabular}
\end{table}

\noindent\textbf{AOPD is essential for effective exploration.}\quad
Removing AOPD while retaining STPR  (``w/o AOPD'') causes CSR to drop by 16.5\% and Acc by 11.8\%.
Crucially, the steepest single-metric decline is on Proact.\ ($-0.76$).
This decrease confirms that AOPD provides a \emph{directional} training signal: by distilling the privileged teacher's targeted strategies into the student, it specifically guides the policy toward initiative-taking behaviors rather than merely improving surface fluency.
Further stripping the turn-level reward (``w/o AOPD + Turn-Level'', equivalent to vanilla GRPO) yields an additional 7.0\% CSR and 6.5\% Acc decline, indicating that sparse outcome rewards alone are insufficient to discover effective proactive strategies.

\noindent\textbf{STPR grounds AOPD's proactive drive in context.}\quad
Removing the turn-level reward (``w/o Turn-Level Reward'') lowers CSR by 5.2\% and Acc by 4.1\%, with noticeable drops in Comm.\ ($-0.24$) and Logic ($-0.35$).
Further removing the entire STPR component (``w/o STPR'', i.e.\ AOPD only) widens the gap to $-7.3$\% CSR and $-7.6$\% Acc, while Comm.\ and Logic continue to deteriorate ($-0.40$ and $-0.58$, respectively).
Interestingly, the Proact.\ score \emph{increases} in both variants ($3.14$ and $3.23$ vs.\ $3.11$ in the full model), revealing that AOPD alone drives the policy toward an \emph{overly aggressive} persuasion style: it eagerly pushes proposals and steers toward acceptance, but neglects the user's intermediate reactions and conversational context.
STPR corrects this tendency by supplying fine-grained, per-turn evaluative feedback that grounds each proactive move in the dialogue state, encouraging the agent to balance initiative with attentive, context-aware responses.

\noindent\textbf{The role of information asymmetry.}\quad
To isolate whether the gain of AOPD stems from \emph{privileged information} or merely from the distillation mechanism itself, we replace AOPD with a vanilla OPD variant (``OPD only''): the teacher is upgraded to a stronger Qwen3-32B model but observes only the same dialogue history as the student, without access to the hidden internal persona $\mathcal{P}^{\mathrm{int}}$.
Despite its substantially larger capacity, OPD yields 7.0\% lower CSR and 5.9\% lower Acc than AOPD, with Proact.\ dropping sharply from 3.23 to 2.39 ($-0.84$).
This confirms that the critical ingredient is not teacher capacity but \emph{information asymmetry}---access to the user's latent concerns enables the teacher to demonstrate targeted, concern-aware strategies that a stronger but equally uninformed teacher cannot discover from dialogue alone.

\subsection{Generalization on Different User Simulator}

\begin{figure}[!tbp]
\centering
\includegraphics[width=.82\linewidth]{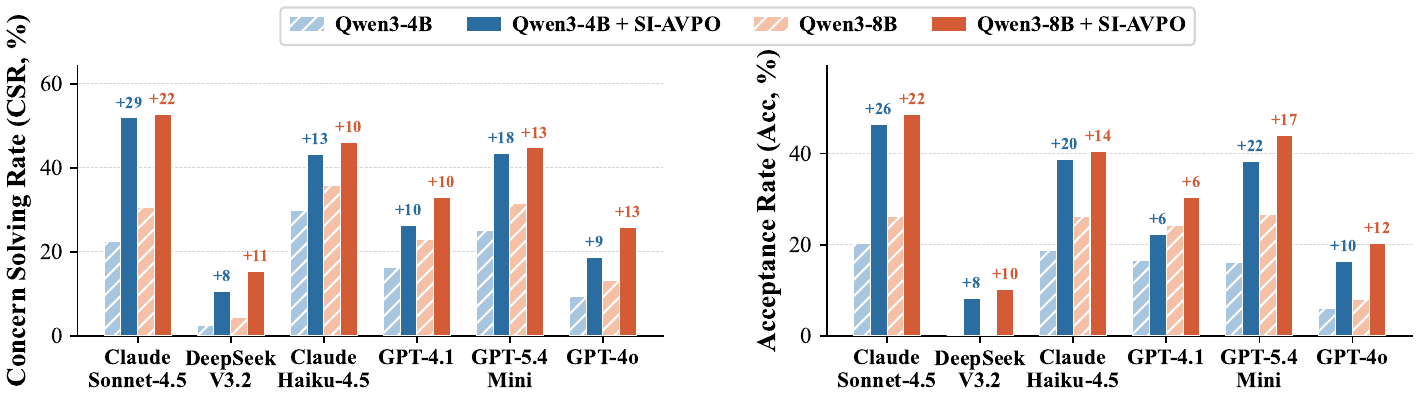}
\caption{Generalization across different user simulators on \textit{Merchant} scene.}
\label{fig:simulator_generalization}
\end{figure}

Figure~\ref{fig:simulator_generalization} evaluates the transferability of the learned dialogue policy to user simulators unseen during training.
An important observation is that different LLMs, even when instantiated with identical persona profiles and instructions, exhibit markedly different levels of strictness in role-playing the user.
For instance, DeepSeek-V3.2 proves to be an exceptionally strict simulator---both Qwen3-4B and Qwen3-8B achieve near-zero Acceptance Rate against it in the base setting.
This variance reflects genuine differences in how each LLM interprets and enacts the user persona rather than differences in task difficulty, as all simulators share the same scenario configuration.
To contextualize these behavioral differences, we also assess simulator fidelity in Section~\ref{sec:simulator_fidelity}, combining topic-diversity analysis with human rationality judgments to measure whether simulator decisions remain grounded in the prescribed persona and dialogue history.

Despite this substantial variation in simulator strictness, \method{} yields consistent improvements across \emph{every} tested simulator.
On average, \method{} improves CSR by 14.8\% and Acc by 15.5\% on Qwen3-4B, and by 13.2\% and 13.8\% on Qwen3-8B, with gains observed on all six simulators without exception.
This uniformity across diverse model families, parameter scales, and conversational styles confirms that \method{} acquires generalizable proactive dialogue competencies rather than overfitting to the idiosyncrasies of a single training-time simulator.

\subsection{User Simulator Fidelity}
\label{sec:simulator_fidelity}

To verify that CUS provides a more faithful training environment than prior external-persona-only simulators~\citep{dai2026sead}, we sample 500 dialogues per simulator and evaluate two aspects: behavioral diversity (\# Unique Topics, Top-3 Topic Share) and Decision Rationality---whether annotators can articulate a grounded reason for the terminal acceptance/rejection decision from the dialogue context. As Table~\ref{tab:simulator_fidelity} shows, CUS expands topic coverage from 8 to 72, reduces Top-3 topic concentration from 56.2\% to 14.4\%, and raises Decision Rationality from 71.4\% to 94.8\% ($\kappa=0.76$). The external-only baseline frequently terminates after content-empty turns (e.g., ``\textit{I clicked / I received it}'') with unmotivated decisions, whereas CUS exposes concrete concerns and willingness shifts that make each decision traceable. These results confirm that the performance gains of \method{} are grounded in optimization against a substantially more faithful user model.

\begin{table}[!htbp]
\centering
\caption{User simulator fidelity on 500 \textit{Merchant} dialogues per simulator. Topic clusters use Qwen3-embedding with annotator verification; Decision Rationality is the majority vote of three blind domain-expert annotators.}
\small
\setlength{\tabcolsep}{6pt}
\begin{tabular}{l ccc}
\toprule
\textbf{Simulator} & \textbf{\# Unique Topics $\uparrow$} & \textbf{Top-3 Topic Share $\downarrow$} & \textbf{Decision Rationality $\uparrow$} \\
\midrule
External-only~\citep{dai2026sead} & 8  & 56.2\% & 71.4\% \\
CUS (Ours)           & 72 & 14.4\% & 94.8\% \\
\bottomrule
\end{tabular}
\label{tab:simulator_fidelity}
\end{table}

\FloatBarrier

\section{Conclusion}

We identified the user's latent concerns as the critical missing variable for proactive task-oriented dialogue and built a framework that models and exploits them.
The Cognitive User Simulator represents each user with hidden internal concerns and synchronous state dynamics, yielding behaviorally diverse interactions whose decisions are grounded in concern progress.
Simulator-Induced Asymmetric-View Policy Optimization converts the resulting privileged state and per-turn transitions into complementary training objectives: asymmetric self-distillation transfers concern-aware behavior into the deployable policy, while state-transition refinement provides fine-grained credit assignment, all without external reward models or judges.
On two real-world food-delivery outbound-call benchmarks, our method matches or surpasses flagship proprietary LLMs and consistently outperforms strong RL baselines, with gains that generalize across diverse unseen user simulators.

\bibliographystyle{unsrtnat}
\bibliography{neurips_2026_ref}




\end{document}